\documentclass{article}

\PassOptionsToPackage{numbers, compress}{natbib}

\usepackage[nonatbib, main, final]{style/neurips_2026}

\usepackage{scrhack} %
\usepackage{algorithm}
\usepackage[noend]{algpseudocode}

\algrenewcommand{\algorithmiccomment}[1]{\hfill {\small \textcolor{darkgray}{$\vartriangleright$ #1}}}
\algrenewcommand\algorithmicindent{2em}
\algrenewcommand\alglinenumber[1]{\small {\textcolor{darkgray}{#1}}}

\makeatletter
\expandafter\patchcmd\csname\string\algorithmic\endcsname{\itemsep\z@}{\itemsep=0.25ex}{}{}
\newcommand\fs@booktabsruled{%
    \def\@fs@cfont{\bfseries\strut}\let\@fs@capt\floatc@ruled
    \def\@fs@pre{\hrule height\heavyrulewidth depth0pt \kern\belowrulesep}%
    \def\@fs@mid{\kern\aboverulesep\hrule height\lightrulewidth\kern\belowrulesep}%
    \def\@fs@post{\kern\aboverulesep\hrule height\heavyrulewidth\relax}%
    \let\@fs@iftopcapt\iftrue
}
\makeatother
\floatstyle{booktabsruled}
\restylefloat{algorithm}

\makeatletter
\newcommand{\mycitestyle}{numeric-comp}
\newcommand{\mybibstyle}{numeric}

\@ifclassloaded{beamer}{
    \renewcommand{\mycitestyle}{alphabetic}
    \renewcommand{\mybibstyle}{alphabetic}

    \preto\fullcite{\AtNextCite{\defcounter{maxnames}{99}}} %
}
\makeatother

\usepackage[
    backend=biber,
    citestyle=\mycitestyle,
    bibstyle=\mybibstyle,
    sorting=none,
    maxcitenames=3,
    maxbibnames=99,
    backref=true,
    natbib=true, %
    giveninits, %
    uniquelist=false,
    ]{biblatex}

\let\cite\parencite

\usepackage{xcolor}

\definecolor{MPLblue}{HTML}{1f77b4}
\definecolor{MPLorange}{HTML}{ff7f0e}
\definecolor{MPLgreen}{HTML}{2ca02c}
\definecolor{MPLred}{HTML}{d62728}
\definecolor{MPLpurple}{HTML}{9467bd}

\definecolor{SNSblue}{rgb}{0.1216, 0.4666, 0.7059}
\definecolor{SNSorange}{rgb}{1.0, 0.4980, 0.0549}
\definecolor{SNSgreen}{rgb}{0.1725, 0.6274, 0.1725}
\definecolor{SNSred}{rgb}{0.84, 0.15, 0.16}
\definecolor{SNSpurple}{rgb}{0.58, 0.40, 0.74}

\definecolor{SNSblue_shaded}{HTML}{8ebad9}
\definecolor{SNSorange_shaded}{HTML}{ffcea3}
\definecolor{SNSgreen_shaded}{HTML}{cae7ca}
\definecolor{SNSred_shaded}{HTML}{ea9293}

\definecolor{SNSblue_colorblind}{HTML}{0173b2}
\definecolor{SNSyellow_colorblind}{HTML}{de8f05}
\definecolor{SNSgreen_colorblind}{HTML}{029e73}
\definecolor{SNSred_colorblind}{HTML}{d55e00}
\definecolor{SNSpink_colorblind}{HTML}{cc78bc}

\definecolor{TUgray}{RGB}{185,184,188}
\definecolor{TUdarkgray}{RGB}{50,65,75}
\definecolor{TUgold}{RGB}{180,160,105}
\definecolor{TUdarkblue}{RGB}{65,90,140}
\definecolor{TUblue}{RGB}{0,105,170}
\definecolor{TUlightblue}{RGB}{80,170,200}
\definecolor{TUlightgreen}{RGB}{130,185,160}
\definecolor{TUgreen}{RGB}{125,165,75}
\definecolor{TUdarkgreen}{RGB}{50,110,30}
\definecolor{TUred}{RGB}{165,30,55}
\definecolor{TUlightred}{RGB}{200,80,60}
\definecolor{TUorange}{RGB}{210,150,0}
\definecolor{TUpurple}{RGB}{175,110,150}
\definecolor{TUocre}{RGB}{200,80,60}
\definecolor{TUviolet}{RGB}{175,110,150}
\definecolor{TUmauve}{RGB}{180,160,150}
\definecolor{TUbeige}{RGB}{215,180,105}
\definecolor{TUbrown}{RGB}{145,105,70}

\definecolor{CUblue}{RGB}{29,79,145}
\definecolor{CUlightblue}{RGB}{185,217,235}
\definecolor{CUdarkblue}{RGB}{0,48,135}
\definecolor{CUgray}{RGB}{117,120,123}
\definecolor{CUlightgray}{RGB}{208,208,206}
\definecolor{CUdarkgray}{RGB}{83,86,90}
\definecolor{CUmagenta}{RGB}{174,37,115}
\definecolor{CUyellow}{RGB}{255,152,0}
\definecolor{CUorange}{RGB}{185,71,0}
\definecolor{CUred}{RGB}{166,10,61}
\definecolor{CUgreen}{RGB}{118,136,29}

\colorlet{maincolor}{CUblue}		        %
\colorlet{secondcolor}{CUblue}		        %

\colorlet{lightgraycolor}{CUlightgray}  %
\colorlet{darkgraycolor}{TUdarkgray}	%

\colorlet{alertcolor}{CUred}
\colorlet{alertlightcolor}{CUred!50!white}

\makeatletter
\@ifclassloaded{beamer}{
    \colorlet{citecolor}{CUblue}
    \colorlet{citelightcolor}{CUlightblue}
    \colorlet{linkcolor}{CUblue}
    \colorlet{linklightcolor}{CUlightblue}
    \colorlet{urlcolor}{CUblue}
    \colorlet{urllightcolor}{CUlightblue}
}{
    \colorlet{citecolor}{MPLblue}
    \colorlet{linkcolor}{black}
    \colorlet{urlcolor}{MPLblue}
}
\makeatother

\usepackage{enumitem}

\usepackage{caption}
\usepackage[labelformat=simple]{subcaption}

\usepackage{graphicx}
\usepackage{pgfplots}
\pgfplotsset{compat=1.15}
\usepackage{adjustbox}
\usepackage{tcolorbox}
\usepackage{tikz}

\DeclareRobustCommand{\colorline}[1]{%
	\begin{tikzpicture}
		\raisebox{1.5pt}{
			\draw[#1,solid,line width=1.5pt] (0,0) -- (1em,0);
		}
	\end{tikzpicture}%
}

\usepackage{amsmath}
\usepackage{amssymb}
\usepackage{mathtools} %
\usepackage{bm}
\usepackage{etoolbox}

\allowdisplaybreaks    %

\DeclarePairedDelimiterX{\set}[1]\{\}{%

#1}

\newcommand*{\R}{\mathbb{R}}

\renewcommand*{\vec}[1]{{\bm{#1}}}

\newcommand*{\mat}[1]{{\bm{#1}}}

\DeclarePairedDelimiterXPP\linspan[1]{\operatorname{span}}{(}{)}{}{#1}
\DeclarePairedDelimiterXPP\im[1]{\operatorname{im}}{(}{)}{}{#1}
\let\ker\relax %
\DeclarePairedDelimiterXPP\ker[1]{\operatorname{ker}}{(}{)}{}{#1}
\DeclarePairedDelimiterXPP\rank[1]{\operatorname{rank}}{(}{)}{}{#1}
\DeclarePairedDelimiterXPP\trace[1]{\operatorname{tr}}{(}{)}{}{#1}
\let\det\relax
\DeclarePairedDelimiterXPP\det[1]{\operatorname{det}}{(}{)}{}{#1}

\newcommand*{\tr}{{\mathsf{T}}}
\renewcommand{\top}{{\tr}} %

\DeclareSymbolFont{stmry}{U}{stmry}{m}{n}
\DeclareMathSymbol\obar\mathrel{stmry}{"3A}
\DeclareMathSymbol\otimes\mathrel{stmry}{"0F}
\DeclareMathSymbol\ominus\mathrel{stmry}{"17}
\makeatletter
\newcommand{\superimpose}[2]{%
  {\ooalign{$#1\@firstoftwo#2$\cr\hfil$#1\@secondoftwo#2$\hfil\cr}}}
\makeatother

\makeatother

\newcommand{\vx}{\vec{x}}
\newcommand{\vy}{\vec{y}}
\newcommand{\vz}{\vec{z}}

\newcommand{\vmu}{\vec{\mu}}

\newcommand{\mI}{\mat{I}}

\newcommand{\mS}{\mat{S}}

\newcommand{\mX}{\mat{X}}

\newcommand{\mId}{\mat{I}}

\DeclarePairedDelimiterX\abs[1]{\lvert}{\rvert}{
  \ifblank{#1}{\:\cdot\:}{#1}
}
\DeclarePairedDelimiterX\norm[1]{\lVert}{\rVert}{
  \ifblank{#1}{\:\cdot\:}{#1}
}
\DeclarePairedDelimiterX\innerprod[2]{\langle}{\rangle}{
  \ifblank{#1}{\:\cdot\:}{#1},\ifblank{#2}{\:\cdot\:}{#2}}

\DeclareDocumentCommand\partialderivative{ s o m g g d() }
{ %
  \IfBooleanTF{#1}
  {\let\fractype\flatfrac}
  {\let\fractype\frac}
  \IfNoValueTF{#4}
  {
    \IfNoValueTF{#6}
    {\fractype{\partial \IfNoValueTF{#2}{}{^{#2}}}{\partial #3\IfNoValueTF{#2}{}{^{#2}}}}
    {\fractype{\partial \IfNoValueTF{#2}{}{^{#2}}}{\partial #3\IfNoValueTF{#2}{}{^{#2}}} \argopen(#6\argclose)}
  }
  {
    \IfNoValueTF{#5}
    {\fractype{\partial \IfNoValueTF{#2}{}{^{#2}} #3}{\partial #4\IfNoValueTF{#2}{}{^{#2}}}}
    {\fractype{\partial^2 #3}{\partial #4 \partial #5}}
  }
}

\NewDocumentCommand{\jac}{o m m o}{%
  \operatorname{D}\IfNoValueF{#1}{_{#1}}\mathopen{}%
  #2\mathopen{}\left( #3 \right)\mathclose{}%
  \IfNoValueF{#4}{%
    \IfNoValueTF{#1}{%
      |_{#3}
    }{%
      |_{#1=#4}
    }
  }
}

\NewDocumentCommand{\grad}{o m m o}{%
  \nabla\IfNoValueF{#1}{_{#1}}\mathopen{}%
  #2\mathopen{}\left( #3 \right)\mathclose{}%
  \IfNoValueF{#4}{%
    \IfNoValueTF{#1}{%
      |_{#3}
    }{%
      |_{#1=#4}
    }
  }
}

\NewDocumentCommand{\natgrad}{o m m o}{%
  \tilde{\nabla}\IfNoValueF{#1}{_{#1}}\mathopen{}%
  #2\mathopen{}\left( #3 \right)\mathclose{}%
  \IfNoValueF{#4}{%
    \IfNoValueTF{#1}{%
      |_{#3}
    }{%
      |_{#1=#4}
    }
  }
}

\NewDocumentCommand{\hessian}{o m m o}{%
  \nabla^2\IfNoValueF{#1}{_{#1}}\mathopen{}%
  #2\mathopen{}\left( #3 \right)\mathclose{}%
  \IfNoValueF{#4}{%
    \IfNoValueTF{#1}{%
      |_{#3}
    }{%
      |_{#1=#4}
    }
  }
}

\makeatletter
\newcommand*{\@probsymbol}{\mathrm{P}}
\newcommand*{\@given}[1]{%
  \nonscript\:#1\vert
  \allowbreak
  \nonscript\:
  \mathopen{}}

\providecommand*{\given}{}
\DeclarePairedDelimiterXPP{\@prob}[1]{\@probsymbol}{(}{)}{}{%
  \renewcommand*{\given}{\@given{\delimsize}}%
  #1}
\newcommand*{\prob}[1]{\ifblank{#1}{\@probsymbol}{\@prob*{#1}}}
\makeatother

\NewDocumentCommand{\expval}{o o m}{%
  \operatorname{\mathbb{E}}\IfNoValueF{#1}{_{#1\IfNoValueF{#2}{\sim #2}}}\mathopen{}\left( #3 \right)
}
\NewDocumentCommand{\cov}{o o m}{%
  \operatorname{Cov}\IfNoValueF{#1}{_{#1\IfNoValueF{#2}{\sim #2}}}\mathopen{}\left( #3 \right)
}
\NewDocumentCommand{\var}{o o m}{%
  \operatorname{Var}\IfNoValueF{#1}{_{#1\IfNoValueF{#2}{\sim #2}}}\mathopen{}\left( #3 \right)
}

\newcommand*{\gaussianpdf}[3]{%
  {\ensuremath{\operatorname{\mathcal{N}}\mathopen{}\left(#1; #2, #3\right)}}%
}

\NewDocumentCommand{\entropy}{o o m}{%
\operatorname{H}\IfNoValueF{#1}{_{#1\IfNoValueF{#2}{\sim #2}}}\mathopen{}\left( #3 \right)
}
\NewDocumentCommand{\infogain}{o o m}{%
\operatorname{IG}\IfNoValueF{#1}{_{#1\IfNoValueF{#2}{\sim #2}}}\mathopen{}\left( #3 \right)
}

\NewDocumentCommand{\dw}{O{2} O{} m m}{%
  \operatorname{W}\IfNoValueF{#1}{_{#1}}\IfNoValueF{#2}{^{#2}}\mathopen{}\left( \ifblank{#3}{\:\cdot\:}{#3},\ifblank{#4}{\:\cdot\:}{#4} \right)
}

\DeclareMathOperator*{\argmax}{arg\,max}

\DeclarePairedDelimiterXPP\bigO[1]{\mathcal{O}}{(}{)}{}{#1}
\DeclarePairedDelimiterXPP\smallo[1]{o}{(}{)}{}{#1}
\DeclarePairedDelimiterXPP\bigOmega[1]{\Omega}{(}{)}{}{#1}
\DeclarePairedDelimiterXPP\smallomega[1]{\omega}{(}{)}{}{#1}
\DeclarePairedDelimiterXPP\bigTheta[1]{\Theta}{(}{)}{}{#1}

\newcommand*{\lr}{\eta}
\newcommand*{\momentum}{\gamma}

\newcommand*{\inputdim}{D_{\text{in}}}

\newcommand*{\numclasses}{C}
\newcommand*{\numtraindata}{N}
\newcommand*{\traininputs}{\mat{X}}

\newcommand*{\traintargets}{\vec{y}}

\newcommand*{\symboltestdata}{{\mathrm{test}}}
\newcommand*{\numtestdata}{{N_\symboltestdata}}

\newcommand*{\testpoint}{\vec{x}_\symboltestdata}

\newcommand*{\batchsize}{\numtraindata_b}

\newcommand*{\params}{\vec{w}}

\newcommand*{\paramspacedim}{P}

\newcommand*{\loss}{\ell}

\newcommand*{\idxlayer}{l}

\newcommand*{\variationalparams}{\vec{\theta}}

\newcommand*{\meanparam}{\vmu}
\newcommand*{\covfactorparam}{\mS}

\newcommand*{\covrank}{R}
\newcommand*{\variationalpdf}{q}
\newcommand*{\expectedloss}{\bar{\loss}}

\newcommand*{\idxdata}{n}

\newcommand*{\idxtime}{t}
\newcommand*{\idxoptimstep}{t}

\NewDocumentCommand{\settime}{O{\idxtime}}{_{#1}}
\NewDocumentCommand{\setlayer}{O{\idxlayer}O{}}{^{(#1)#2}}

\makeatletter
\@ifclassloaded{beamer}{
  \usepackage{pdfpages}

  \hypersetup{
    pdfkeywords={SP-Right}
  }
}
\makeatother

\usepackage{titletoc}

\makeatletter
\@ifclassloaded{beamer}{}{
    \usepackage{tocloft}
    \setlength\cftbeforesecskip{0em}
    \setlength\cftbeforesubsecskip{-0.5em}
    \setlength\cftbeforesubsubsecskip{-0.5em}
    
}
\makeatother

\usepackage[numbered,open=true]{bookmark}

\usepackage{booktabs}
\usepackage{longtable}
\usepackage{multirow}
\usepackage{rotating}
\usepackage{siunitx}

\usepackage[normalem]{ulem}
\usepackage{xspace}

\newcommand*{\ie}{i.e.\@\xspace}

\usepackage{amsthm}
\usepackage{thmtools}
\usepackage{thm-restate}

\declaretheoremstyle[
    headfont=\normalfont\bfseries,
    notefont=\normalfont,
    bodyfont=\normalfont,
    headpunct={},
    postheadspace=\newline,
]{definitionstyle}

\declaretheoremstyle[
    headfont=\normalfont\bfseries,
    notefont=\normalfont,
    bodyfont=\normalfont\itshape,
    headpunct={},
    postheadspace=\newline,
]{lemmastyle}

\declaretheoremstyle[
    headfont=\normalfont\bfseries,
    notefont=\normalfont,
    bodyfont=\normalfont\itshape,
    headpunct={},
    postheadspace=\newline,
]{theoremstyle}

\declaretheoremstyle[
    headfont=\normalfont\bfseries,
    notefont=\normalfont,
    bodyfont=\normalfont,
    headpunct={},
    postheadspace=\newline,
]{remarkstyle}

\makeatletter
\@ifclassloaded{beamer}{
    
}{

    \declaretheorem[style=remarkstyle,name=Remark]{remark}

}
\makeatother

\usepackage[
    textsize=scriptsize,
    figwidth=0.99\linewidth,
]{todonotes}

\usepackage{hyperref}
\hypersetup{
    unicode,                    %
    pdfborder       = {0 0 0},  %
    bookmarksdepth  = subsection,
    bookmarksopen   = true,     %
    linktoc         = all,      %
    breaklinks      = true,
    colorlinks      = true,
    linkcolor       = linkcolor,
    citecolor       = citecolor,
    urlcolor        = urlcolor,
}

\usepackage[
    capitalize,
    nameinlink,
]{cleveref}

\usepackage[utf8]{inputenc} %
\usepackage[T1]{fontenc}    %
\usepackage{amsfonts}       %
\usepackage{nicefrac}       %
\usepackage{microtype}      %
\usepackage{xcolor}         %
\definecolor{C0}{HTML}{1F77B4}  %
\definecolor{C1}{HTML}{FF7F0E}  %

\title{Characterizing the Edge of Stability\\ in Variational Training Without Priors}

\author{%
  Juraj Marusic\thanks{Equal contribution.} \\
  Columbia University \\
  \And
  Jonathan Wenger\footnotemark[2] \\
  Columbia University \\
  \And
  Beau Coker \\
  Columbia University \\
  \And
  John P.~Cunningham \\
  Columbia University \\
}

\begin{document}

\setcounter{footnote}{1}
\maketitle
\setcounter{footnote}{0}

\begin{abstract}
    Variational approaches in deep learning promise to deliver improved uncertainty quantification and out-of-distribution generalization by inferring a distribution over weights, regularized by a divergence to a chosen prior.
    However, prior elicitation presents a significant challenge in deep learning, resulting in the common practice of significantly downweighting the regularization term in the variational objective.
    Recently, it was proposed to train solely via the expected loss by relying on implicit regularization from the optimizer, rather than explicit regularization from the divergence to the prior.
    While this approach largely sidesteps the issue of prior elicitation, current theory lacks an understanding of how key optimization hyperparameters impact the implicit regularization.
    Here, we theoretically and empirically demonstrate that larger learning rates and \emph{fewer} parameter samples lead to flatter minima of the loss landscape, which are known to generalize better. 
    We show that training on the expected loss operates at a modified edge of stability, characterized by the signal-to-noise ratio of the gradient estimates and the number of Monte Carlo samples. 
    Experiments on out-of-distribution benchmark datasets confirm the practical significance of the theoretical results. 
\end{abstract}

\section{Introduction}
There has been a growing interest in studying Bayesian methods in deep learning to improve out-of-distribution generalization and uncertainty quantification \cite{papamarkouPositionBayesianDeep2024}. 
While exact Bayesian inference can perform well \cite{Izmailov2021WhatAre}, its computational demands require approximate inference techniques, which in combination with challenges in prior elicitation have been shown to lead to undesired inductive biases like severe underfitting \cite{Foong2020ExpressivenessApproximate,cinquinPathologiesPriorsInference2021,cokerWideMeanFieldBayesian2022}.
In variational approaches, this is commonly addressed by significantly ``tempering'', i.e. downweighting the divergence to the prior in the loss function \cite{Wenzel2020HowGood}, or even omitting this explicit regularization term from the objective entirely, as proposed recently by \citet{wengerVariationalDeepLearning2026}. This results in an approach which relies entirely on implicit regularization imposed by the optimization procedure and its hyperparameters rather than on explicit regularization from the prior.

Implicit regularization has been identified as a key explanation for the generalization performance of modern deep learning models. 
In deep learning, where there are often multiple minima of the training loss, the curvature of the loss at (global) minima plays a central role in determining how the corresponding model will behave on test data.
Flatter minima (i.e., less sharp, as measured by the largest eigenvalue of the Hessian of the loss) correspond to more ``regular'' functions \cite{Mulayoff2021ImplicitBias,Dherin2022WhyNeural,Nacson2023ImplicitBias} and there is ample evidence that they generalize better \cite{Hochreiter1997FlatMinima,Hinton1993KeepingNeural,Dziugaite2017ComputingNonvacuous,Keskar2017LargeBatchTraining,Foret2021SharpnessAwareMinimization}.
This phenomenon, known as the \emph{edge of stability} (EoS), whereby the sharpness of deep neural networks trained with gradient descent (GD) at learning rate $\lr$ will progressively increase before oscillating near \( 2/\eta \), provides a link between the optimization procedure and the sharpness \cite{wuHowSGDSelects2018, jastrzebskiBreakEven2020, cohenGradientDescentNeural2020}.\footnote{For the cross-entropy loss sharpness decays asymptotically to zero after reaching the edge of stability \cite{cohenGradientDescentNeural2020}.}

One way to guide optimization towards flatter minima is to inject noise while training, most commonly through minibatching in stochastic gradient descent (SGD) \cite{zhuAnisotropicNoiseStochastic2019, zhangWhichAlgorithmicChoices2019, nguyenFirstExitTime2019, WuImplicitStability2023}. 
Recent work by \citet{Ghosh2025VariationalLearningFindsFlatter} has shown that adding Gaussian noise to the parameters during training decreases the upper bound on stable learning rates, thus allowing for even flatter minima to be obtained than with SGD only.
This ``weight-perturbed'' training approach is a form of variational inference with a fixed, diagonal covariance structure and where only the mean parameters are used at test time.
Unfortunately, the fixed amount of noise added throughout optimization can induce biases that hurt in-distribution performance, as it may over-regularize the learning problem.
Hence, the magnitude of the injected noise serves as a trade-off between in- and out-of-distribution performance.
By instead training the injected noise distribution with the unregularized variational objective, relying solely on the inductive biases of the optimizer, one can often achieve strong results on both types of tasks \cite{wengerVariationalDeepLearning2026}. 
However, training the injected noise distribution introduces new challenges. There are now two learning rates to consider, one for the weight parameters and one for the noise parameters, and it is not altogether clear that training the noise will not undo the impact on sharpness, since the lack of explicit regularization implies that predictive uncertainty will collapse on training data.

\begin{figure*}[t]
    \centering
    \begin{subfigure}[t]{0.49\textwidth}
        \centering
        \includegraphics[width=\textwidth]{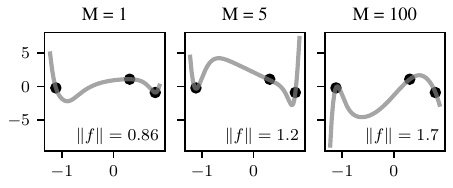}
        \caption{Sharpness \( \geq 1 + 2 \| f \| \) }
        \label{fig:figure1left}
    \end{subfigure}
    \begin{subfigure}[t]{0.49\textwidth}
        \centering
        \includegraphics[width=\textwidth]{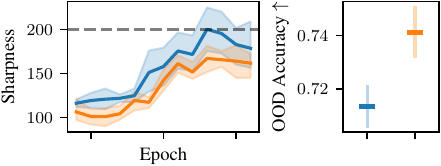}
        \caption{\, \colorline{C0} Standard \quad \colorline{C1} Variational Training }
        \label{fig:figure1right}
    \end{subfigure}
    \caption{\textit{Variational training with \emph{fewer} parameter samples leads to better generalization.}
    The number of parameter samples used to estimate the expected loss in variational training (without prior regularization) affects the sharpness of the minima at convergence. A smaller number of samples \( M \) leads to convergence to flatter minima, which correspond to more regular functions (see (a); c.f. \cite{Mulayoff2021ImplicitBias}), thus explaining the observed improved generalization.
    In (b), we show the sharpness of the loss during standard and variational training of a ResNet on CIFAR-10. Notably, in the variational case the sharpness remains below the standard threshold of \(2 / \text{learning rate}\). When evaluated on a test set with distribution-shift, the variational model generalizes better out-of-distribution.}
    \vspace{-0.5em} %
    \label{fig:figure1}
\end{figure*}

In this work, we study the edge of stability phenomenon in the context of variational deep learning without explicit prior regularization \cite{wengerVariationalDeepLearning2026}. 
We show that training a variational model on a Monte-Carlo estimate of the expected loss, i.e. where the expectation is over injected noise (or, equivalently, parameters by the reparameterization trick), leads to a modified edge of stability regime. 
The sharpness threshold is modified by the signal-to-noise ratio of the gradient estimates as well as the number of samples used to estimate the expected loss. 
As is common in the EoS literature, we prove results for a linear regression model (i.e., a quadratic objective) and then verify the stability conditions on deep networks.
Interestingly, using \emph{fewer} samples to approximate the expected loss results in convergence to flatter minima and thus better generalization. 
This shows that it is the approximation from sampling---often seen as a necessary sacrifice to scale variational deep learning to large models---that is critical to the regularization benefits, and not only expected loss itself or the explicit regularization of the KL divergence. 
\Cref{fig:figure1left} illustrates the findings in a toy model, where using fewer parameter samples results in convergence to flatter minima, which correspond to functions with higher regularity.
\Cref{fig:figure1right} shows that peak sharpness of minibatched variational training remains lower than for standard training, which results in better out-of-distribution generalization.

\paragraph{Contributions.} 
We characterize the implicit regularization properties of variational training via the expected loss. 
Specifically, in the context of variational linear regression, we show that the maximal learning rates of the mean and covariance parameters that guarantee descent are each modified by the signal-to-noise ratio of the gradient estimates and the number of parameter samples used for gradient estimates (\Cref{thm:descent-mc-expected-loss}). 
We extend this result to arbitrary $\beta$-smooth loss landscapes with Gaussian variational families (\Cref{thm:descent-general-loss}). 
In the context of variational deep learning, these results correspond to a modified edge of stability.
We confirm our theoretical results empirically on a set of robustness benchmarks, where approximating the expected loss with just a single sample leads to better in- and out-of-distribution generalization at minimal computational overhead.

\section{Background}
\label{sec:background}

In supervised learning, we are given a training dataset $(\mX, \vy) = \{(\vx_n, y_n)\}_{n=1}^\numtraindata$ and we want to learn a model $f_\params(\vx)$ with parameters $\params$ that maps inputs $\vx$ to outputs $y$. This is usually done by minimizing the empirical loss given by 
\begin{equation} \label{eq:empirical_loss}
    \textstyle %
    \loss (\params) := \loss_0 (f_\params (\mX), \vy) = \sum_{n=1}^\numtraindata \loss_0 (f_\params (\vx_n), y_n),
\end{equation}
where $\loss_0(\cdot, \cdot)$ is a pointwise loss function. 

In order to minimize the empirical loss, we usually resort to gradient-based optimization methods, such as gradient descent. These methods iteratively update the model parameters $\params$ such that:
\begin{equation} \label{eq:gd_update}
    \params_{t + 1} = \params_t - \lr \grad{\loss}{\params_t}.
\end{equation}
The conditions for guaranteed descent in this setting are as follows.
\begin{restatable}[Descent on $\beta$-smooth losses]{lemma}{lemmaDescentStandardLoss} \label{lem:descent-standard-loss}
    Consider the empirical loss from \cref{eq:empirical_loss}. Consider gradient descent updates with learning rate $\eta$ given in \cref{eq:gd_update}. Let $\beta$ denote the smoothness of the loss $\loss$, that is, $ \beta = \sup_\params \| \nabla^2 \loss (\params) \|$. Then,
    \[
    \loss (\params_{t + 1} ) - \loss (\params_t) \leq 0, \quad \text{if} \quad \eta < \frac{2}{\beta}.
    \]
\end{restatable}

While the above lemma provides conditions for the guaranteed descent on smooth functions, deep learning loss landscapes do not satisfy this smoothness assumption. During the training of deep models via gradient descent, as in \cref{eq:gd_update}, the sharpness of the loss, defined as the largest eigenvalue of the Hessian of the loss $\lambda_\text{max} (\nabla^2 \loss (\params))$, tends to increase until it reaches the threshold value of $2 / \lr$. This observation was termed \textit{edge of stability}, and has since been an active area of research. The main implications being that the sharpness of the loss minima can be connected to the properties of the resulting function \cite{Mulayoff2021ImplicitBias,Nacson2023ImplicitBias} and robustness to input perturbations \cite{Dherin2022WhyNeural}.
We now briefly turn to reviewing variational inference and its implicit bias variant.

\paragraph{Variational Inference (VI).} In variational inference, we place a prior distribution $p(\params)$ over the model parameters and approximate the posterior distribution $p(\params | \mX, \vy)$ with a variational distribution $q_\variationalparams(\params)$ with variational parameters $\variationalparams$. The variational parameters are usually learned by minimizing the negative evidence lower bound (ELBO)
\begin{equation*}
    \mathcal{L} (\variationalparams) 
    :=
    \expval[\params][q_\variationalparams(\params)]{- \log p(\vy | \mX, \params)} + \text{KL}({q_\variationalparams},{p}),
\end{equation*}
where $\text{KL}({q_\variationalparams},{p})$ denotes the Kullback-Leibler divergence between the variational distribution and the prior. The first term in the ELBO corresponds to the expected negative log likelihood over the variational distribution, while the second term acts as a regularizer that encourages the variational distribution to stay close to the prior.
\paragraph{Unregularized Variational Training.} 
As the name suggests, in this method, the variational parameters are learned by maximizing the expected loss only (for instance, expected log-likelihood), without the KL regularization term \cite{wengerVariationalDeepLearning2026}. This leads to the minimization of the \emph{expected loss}:
\begin{equation} \label{eq:expected_loss}
\begin{split}
    \expectedloss (\variationalparams) 
    &:=
    \expval[\params][\variationalpdf_\variationalparams(\params)]{\loss_0 (f_\params (\mX), \vy)} 
    =
    \textstyle %
    \sum_{n=1}^\numtraindata \expval[\params][\variationalpdf_\variationalparams(\params)]{\loss_0 (f_\params (\vx_n), y_n)}.
\end{split}
\end{equation}
By minimizing the expected loss, the optimization procedure implictly regularizers variational parameters. In the case of overparametrized linear regression, it can be shown that (S)GD iterates converge to the variational distribution that minimizes the 2-Wasserstein distance to $\variationalpdf_{\variationalparams_0}$ (which can be thought as a ``prior'') among all distributions that achieve zero expected loss \cite{wengerVariationalDeepLearning2026}.

\section{Characterizing the Edge of Stability in Variational Training Without Priors}
\label{sec:benefits-of-sampling}

Throughout this section, we consider variational models with a Gaussian variational family, i.e. $q_\variationalparams(\params) = N(\meanparam, \covfactorparam \covfactorparam^\top)$, where $\variationalparams = (\meanparam, \covfactorparam)$ denotes the variational parameters consisting of mean parameter $\meanparam \in \R^P$ and covariance factor parameter $\covfactorparam \in \R^{P \times R}$ with $R \leq P$. Therefore, the expected loss from \cref{eq:expected_loss} becomes
\begin{equation} \label{eq:gaussian-expected-loss}
   \expectedloss(\variationalparams)  
    =
    \textstyle %
    \sum_{n=1}^\numtraindata \expval[\params][N(\meanparam, \covfactorparam \covfactorparam^\top)]{\loss_0 (f_\params (\vx_n), y_n)}.
\end{equation}

Note that in practice, calculating the above expectation is usually computationally not feasible, and we resort to stochastic approximation using $M$ parameter samples from the variational distribution:
\begin{equation*}
    \expectedloss(\variationalparams) \ \textcolor{SNSred}{\approx}\ \hat{\expectedloss}(\variationalparams) 
    =
    \textstyle %
    \frac{1}{M} \sum_{m=1}^M \sum_{n=1}^\numtraindata \loss_0 (f_{\params_m} (\vx_n), y_n),
\end{equation*}
where $\params_m \sim N(\meanparam, \covfactorparam \covfactorparam^\top)$ are samples from the variational distribution. The mean and covariance factor parameters are then learned via gradient descent on the above stochastic loss. Our goal is to answer the following two questions:
\begin{itemize}
    \item[\textbf{[Q1]}] When parameters are sampled to estimate gradients, what is the largest learning rate for which the expected loss is still guaranteed to decrease on average at every step?
    \item[\textbf{[Q2]}] Does this characterization extend to deep variational models, and how does it modify the edge of stability phenomenon?
    \end{itemize}
We start by studying the case of variational linear models.
\subsection{Variational Linear Model}
\begin{figure*}[t]
    \centering
    \begin{subfigure}[t]{0.32\textwidth}
        \centering
        \includegraphics[width=\textwidth]{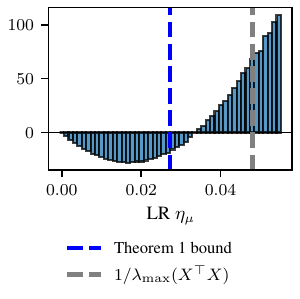}
        \caption{Descent on the mean parameter}
        \label{fig:theory1}
    \end{subfigure}
    \hfill
    \begin{subfigure}[t]{0.32\textwidth}
        \centering
        \includegraphics[width=\textwidth]{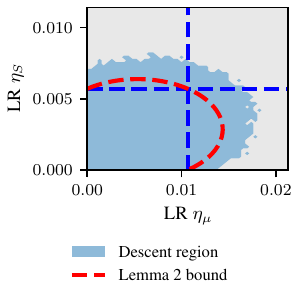}
        \caption{Descent region}
        \label{fig:theory2}
    \end{subfigure}
    \hfill
    \begin{subfigure}[t]{0.32\textwidth}
        \centering
        \includegraphics[width=\textwidth]{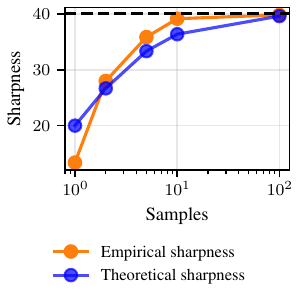}
        \caption{One-layer variational SiLU net}
        \label{fig:theory3}
    \end{subfigure}
    \caption{\textit{Empirical verification of theoretical findings.}
    While \Cref{thm:descent-mc-expected-loss} provides upper bounds on the learning rates, here we demonstrate that these bounds are indeed sharp. In (a), we plot the difference \(\expval[]{\expectedloss(\variationalparams_{t+1})} - \expectedloss(\variationalparams_t)\) after a gradient update of the mean parameter \(\meanparam_t\), while keeping the covariance parameter \(\covfactorparam\) fixed. We observe that the theoretical bound tightly characterizes the region in which descent is guaranteed. Furthermore, we see that an adjustment to the learning rate is necessary—the standard bound from the non-variational setting, \(1 / \lambda_{\max}(\mX^\top \mX)\), is not valid here. In (b), we compare the joint conditions on the learning rates for both the mean and covariance parameters, again confirming their tightness. Finally, in (c) we illustrate how these results translate to neural networks by comparing the sharpness of the minima with respect to the mean parameter under unregularized variational optimization of a one-layer SiLU network to the approximate theoretical sharpness, given by the mapping
    \(M \mapsto 2 / \lr \cdot 1 / (1 + 1/M)\).}
    
    \label{fig:theory_figures}
\end{figure*}

Consider the variational linear model $f_\params(\vx) = \vx^\top \params$, with $\params \sim N(\meanparam, \covfactorparam \covfactorparam^\top)$. The expected loss is given by
\begin{equation} \label{eq:variational-linear-model-expected-loss}
\begin{split}
    \expectedloss(\variationalparams) 
    &=
    \expval[\params][N(\meanparam, \covfactorparam \covfactorparam^\top)]{\| \mX \params - \vy \|_2^2} 
    =
    \| \mX \meanparam - \vy \|_2^2 + \| \mX \covfactorparam \|_F^2,
\end{split}
\end{equation}
and corresponding Monte Carlo-approximated loss is given by
\begin{equation} \label{eq:variational-linear-model-mc-expected-loss}
    \hat{\expectedloss}(\variationalparams) 
    =
    \textstyle %
    \frac{1}{M} \sum_{m=1}^M \| \mX (\meanparam + \covfactorparam \vz_m) - \vy \|_2^2,
\end{equation}
where $\vz_m \sim N(0, \mI)$ are independent normal random variables.
We are now ready to state the main result of this section, which characterizes the maximal learning rates that guarantee descent of the expected loss when using sampled gradients of the Monte Carlo objective defined in \cref{eq:variational-linear-model-mc-expected-loss}.
\begin{restatable}[Descent via sampled gradients]{theorem}{myfavtheorem} \label{thm:descent-mc-expected-loss}
	Consider the gradient descent updates for the variational linear model in \cref{eq:variational-linear-model-expected-loss}, where the gradients are estimated via the Monte Carlo objective in \cref{eq:variational-linear-model-mc-expected-loss}. 
    Specifically, let
    \[
        \begin{pmatrix}
            \meanparam_{t+1} \\
            \covfactorparam_{t+1}
        \end{pmatrix}
        =
        \begin{pmatrix}
            \meanparam_t \\
            \covfactorparam_t
        \end{pmatrix}
        -
        \begin{pmatrix}
            \lr_{\meanparam} & 0 \\
            0 & \lr_{\covfactorparam}
        \end{pmatrix}
        \nabla_{(\meanparam,\covfactorparam)}
        \hat{\expectedloss}(\meanparam_t,\covfactorparam_t).
    \] 
    Then the expected loss is non-increasing in expectation over the Monte Carlo samples \(\{z_m\}_{m=1}^M\), i.e.,
    \( \expval[]{\expectedloss (\variationalparams_{t + 1} )} - \expectedloss (\variationalparams_t) \leq 0, \)
    provided that the learning rates satisfy
    \[
        \lr_\meanparam 
        <
        \frac{1}{\lambda_\text{max} (\mX^\top \mX)} \cdot \frac{1}{1 + \frac{1}{M} \frac{ \| \mX^\top \mX \covfactorparam \|_F^2}{\| \mX^\top (\mX \meanparam - \vy) \|_2^2}},
    \]
    and
    \[
        \lr_\covfactorparam
        <
        \frac{1}{\lambda_\text{max} (\mX^\top \mX)} \cdot \frac{1}{ 1 + \frac{1}{M} \left ( 1 + R + \frac{\| \mX^\top (\mX \meanparam - \vy) \|_2^2}{\tfrac{1}{R}\| \mX^\top \mX \covfactorparam \|_F^2} \right )}.
    \]
\end{restatable}
We now compare this result to the standard setting described in \cref{lem:descent-standard-loss}. For standard linear regression, the smoothness constant of the loss in \cref{eq:empirical_loss} is given by \( \beta = 2 \lambda_{\max}(\mX^\top \mX) \). Consequently, the maximal learning rate that guarantees loss descent in this setting satisfies \( \eta_\params < 1 / \lambda_{\max}(\mX^\top \mX) \). In contrast, comparing this bound to the maximal learning rate for the mean parameters in \cref{thm:descent-mc-expected-loss}, we observe that the admissible learning rate is always smaller when using sampled gradients for expected loss descent.
Intuitively, this difference makes sense, as the stochasticity from minibatch sampling introduces additional noise to the gradient estimates, requiring smaller learning rates for guaranteed expected loss descent on average. In \Cref{fig:theory1}, we empirically verify \cref{thm:descent-mc-expected-loss} by plotting the difference $\expval[]{\expectedloss (\variationalparams_{t + 1} )} - \expectedloss (\variationalparams_t)$ as a function of the mean parameter learning rate $\eta_\meanparam$, without the covariance update. We see that the theoretical threshold from \Cref{thm:descent-mc-expected-loss} provides a sharp necessary condition for guaranteed expected loss descent. 

Furthermore, we see that as the number of samples $M$ increases, the maximal learning rates approach the standard setting, achieving the standard maximal learning rates in the limit $M \to \infty$. In order to explain the ratio \( \| \mX^\top \mX \covfactorparam \|_F^2 / \| \mX^\top (\mX \meanparam - \vy) \|_2^2 \), note that it can be interpreted as the signal-to-noise ratio of the single sample estimate of the mean parameter gradient. More precisely, let \( \hat{g} = \nabla_{\meanparam} \| \mX (\meanparam + \covfactorparam z) - \vy \|_2^2 \) be the single sample gradient estimate of the mean parameter, we have that its covariance (noise) is given by
\begin{equation*}
\begin{split}
    \var[]{\hat{g}} 
    &=
    4 \mX^\top \mX \covfactorparam \covfactorparam^\top \mX^\top \mX
    \implies
    \trace{\var[]{\hat{g}}}
    =
    4 \| \mX^\top \mX \covfactorparam \|_F^2.
\end{split}
\end{equation*}
At the same time, the mean (signal) is given by
\begin{equation*}
    \begin{split}
    \expval[]{\hat{g}}
    &=
    2 \mX^\top (\mX \meanparam - \vy)
    \implies
    \| \expval{\hat{g}} \|_2^2
    =
    4 \| \mX^\top (\mX \meanparam - \vy) \|_2^2.
\end{split}
\end{equation*}
Hence, the larger signal-to-noise ratio (SNR) of the gradient estimates, the closer the maximal learning rate is to that of the standard setting. A similar interpretation holds for the learning rate of the covariance factor parameter. As we will see, an analogous analysis based on the SNR extends to arbitrary models with smooth losses (see \Cref{thm:descent-general-loss}).

In \Cref{fig:theory2}, we show the empirical verification of \Cref{thm:descent-mc-expected-loss}. We plot the indicator $\mathbf{1} \{ \expval[]{\expectedloss (\variationalparams_{t + 1} )} - \expectedloss (\variationalparams_t) \geq 0 \}$ as a function of learning rates $\eta_\meanparam, \eta_\covfactorparam$. We see that the theoretical thresholds from \Cref{thm:descent-mc-expected-loss} provide a necessary condition for guaranteed expected loss descent. The provided theoretical area has a rectangular shape, and does not take into the account the joint effect of both learning rates. To that end, we provide the following result that characterizes joint conditions on the learning rates for guaranteed expected loss descent. As can be seen from the same figure, the ellipsoid characterization provides a tighter lower bound on the expected loss descent area.

\begin{restatable}[Joint conditions on the learning rates]{lemma}{lemmaJointDescent}
    Consider the setup of \cref{thm:descent-mc-expected-loss}. Define 
    \[
         \gamma = \frac{\| \mX^\top (\mX \meanparam - \vy) \|_2^2}{ \| \mX^\top \mX \covfactorparam \|_F^2}
    \]
    to be the signal-to-noise ratio of the one sample mean parameter gradient estimate.
    Then
    \( \expval[]{\expectedloss (\variationalparams_{t + 1} )} - \expectedloss (\variationalparams_t) \leq 0 \)
    provided that
    \begin{equation*}
        \gamma \left ( a_\mu \lr_\meanparam^2 - \lr_\meanparam \right ) + \left ( a_S \lr_\covfactorparam^2 - \lr_\covfactorparam \right ) < 0,
    \end{equation*}
    which defines an elliptical stability region in the \( (\lr_\meanparam, \lr_\covfactorparam) \) plane. Here,
    \[
        a_\mu = \lambda_\text{max} (\mX^\top \mX) \left ( 1 + \frac{1}{M \gamma} \right ),
        \quad
        \text{and}
        \quad
        a_S = \lambda_\text{max} (\mX^\top \mX) \left ( \frac{R + M + 1}{M} + \frac{R \gamma}{M}  \right ).
    \]
\end{restatable}
To further understand the SNR, assume for a moment that we are directly optimizing the expected loss in \cref{eq:variational-linear-model-expected-loss} via gradient descent. \citet{wengerVariationalDeepLearning2026} have shown that, assuming the data is coming from the prior predictive, and if further assuming that \( \| \mX^\top (\mX \meanparam_0 - \vy) \|_2^2 =  \| \mX^\top \mX \covfactorparam_0 \|_F^2 \), we have that these two terms are equal throughout the optimization, i.e. \( \| \mX^\top (\mX \meanparam_t - \vy) \|_2^2 =  \| \mX^\top \mX \covfactorparam_t \|_F^2 \) for all $t$ (see \Cref{thm:nonasymptotic-error-sgd} in Supplementary material). In this case, the SNR $\gamma$ is equal to $1$. Note that this need not be the case when using sampled gradients, as the stochasticity from sampling can break this equality. Nevertheless, this provides some intuition on the typical values of $\gamma$ during the optimization. With this in mind, the learning rate when using sampled gradients (\Cref{thm:descent-mc-expected-loss}) should be smaller compared to the standard setting by a factor of approximately \( (1 + 1 / M) \) for the mean parameters. We will see later that this approximately holds in deep learning settings as well.

\begin{remark}[Differences from noise induced by SGD]
    It is instructive to contrast the Monte Carlo noise in \cref{eq:variational-linear-model-mc-expected-loss} with the minibatch noise in standard SGD on the squared loss. For minibatch SGD, the per-sample gradient variance scales with the residual $\mX \meanparam - \vy$. In the interpolating regime it therefore vanishes at the optimum. The noise is intrinsic to the data and the residual. In contrast, the Monte Carlo noise here is governed by $\| \mX^\top \mX \covfactorparam \|_F^2$ and arises from sampling the parameters $\params \sim N(\meanparam, \covfactorparam \covfactorparam^\top)$ rather than from the data. It persists at $\mX \meanparam = \vy$ and vanishes only in the degenerate limit $\covfactorparam \to 0$.
\end{remark}

\subsection{Extensions Beyond Linear Models}
We can extend the results from the linear model to arbitrary models with smooth losses. Consider the Gaussian variational model and the expected loss from \cref{eq:gaussian-expected-loss}. We have the following result regarding the maximal learning rate on the mean parameter for the guaranteed expected loss descent when using minibatch sampled gradients.
\begin{restatable}[Extension to smooth losses]{theorem}{theoremGeneralLoss} \label{thm:descent-general-loss}
    Consider the variational model and expected loss in \cref{eq:gaussian-expected-loss}. Let \( \beta = \sup_\params \| \nabla^2 \loss (\params) \| \) denote the smoothness constant of the loss $\loss$. Consider gradient descent updates on the mean parameter with learning rate $\eta_\meanparam,$ given by $\meanparam_{t + 1} = \meanparam_t - \eta_\meanparam \grad[\meanparam]{\hat{\expectedloss}}{\meanparam_t}$. Then
    \( \expval[]{\expectedloss (\variationalparams_{t + 1} )} - \expectedloss (\variationalparams_t) \leq 0 \)
    if
    \[
        \lr_\meanparam 
        <
        \frac{2}{\beta} \left ( \frac{1}{ 1 + \frac{1}{M}  \frac{\trace*{ \cov[\params][\variationalpdf_\variationalparams]{\nabla \loss (\params) }}}{\| \expval[\params][\variationalpdf_\variationalparams]{\nabla \loss (\params)} \|_2 ^2}} \right ),
    \]
    where the expectation is taken over the randomness in the Monte Carlo samples $\{ z_m \}_{m=1}^M$.
\end{restatable}
We see that similar to the linear model case, the maximal learning rate for guaranteed expected loss descent is always smaller when using sampled gradients and characterized by the signal-to-noise ratio of the gradient estimates and the number of Monte Carlo samples. As before, when the number of samples $M$ increases, the maximal learning rate approaches the standard setting, achieving it in the limit $M \to \infty$.

As in the standard optimization setting, we expect this results to translate to the edge of stability phenomenon when training deep variational models via gradient descent on the expected loss using sampled gradients. Specifically, we expect the sharpness of the loss minima at the mean parameters to oscillate around the threshold value given by
\begin{equation} \label{eq:sharpness-threshold-general-loss}
    \lambda_\text{max} (\nabla^2 \loss (\meanparam_\infty)) 
    \approx
    \frac{2}{\eta} \left ( \frac{1}{ 1 + \frac{1}{M \gamma} }  \right ),
\end{equation}
where, as before, we denote the signal-to-noise ratio of the gradient estimates by \( \gamma \).

Following the reasoning from the linear model case, we can expect the SNR \( \gamma \) to be somehwat stable throughout the training. To that end, we train a one-layer SiLU network on synthetic data using unregularized variational learning with Gaussian variational family. In \Cref{fig:theory3}, we plot the sharpness at the mean parameters at convergence as a function of the number of samples $M$ used for gradient estimates, along with the theoretical approximation, the mapping \( M \mapsto 2 / \eta \cdot (1 / (1 + 1 / M)) \). The two curves considerably match, supporting the claim that the characterization from the linear model case translates to deeper variational models as well. Further empirical analysis is deferred to the Experiments section.

\subsection{Related Work}

Works on implicit regularization can largely be divided into those that consider gradient flow \cite{aliImplicitRegGradientFlow2020, bowmanImplicitBiasMSE2022,boursierGradientFlowDynamics2022, jinImplicitBiasGradient2023,JianfaWideOnR2023,timorRankMinimizationReLU2023} and those that consider (stochastic) gradient descent \cite{gunasekarCharacterizingImplicitBias2018, WuImplicitStability2023,bowmanLinearDiagonal2022,barrettImplicitGradientRegularization2021,smithOriginImplicitRegularization2021}. It has been established that the sharpness (maximum eigenvalue of the Hessian of the training loss) progressively increases before leveling off near \( 2/\eta \), called the \emph{edge of stability} (EoS) regime \cite{cohenGradientDescentNeural2020, cohenAdaptiveEdgeOfStability2024, cohenCentralFlows2024, wuHowSGDSelects2018, jastrzebskiRelationBetweenSharpest2018, jastrzebskiBreakEven2020}. 
There is an abundance of evidence that less sharp (``flatter'') minima correlate with better generalization, both empirical \cite{Keskar2017LargeBatchTraining,wuHowSGDSelects2018,Foret2021SharpnessAwareMinimization} and theoretical \cite{Dziugaite2017ComputingNonvacuous,Mulayoff2021ImplicitBias,Dherin2022WhyNeural,Nacson2023ImplicitBias}. 
Many works consider the regularizing effect of stochasticity in training. 
Perhaps most well-known is from the stochasticity in batching for gradient updates, i.e. SGD. 
It has been shown to have improved performance over GD \cite{WuImplicitStability2023}, with further improvements from smaller batches \cite{marekSmallBatchSizeLLMs2025}. 
Parameters can also be stochastically adjusted during gradient steps. 
Dropout \cite{nitishDropout2014} multiplies parameters by Bernoulli noise while ``weight-perturbed'' inference \cite{zhuAnisotropicNoiseStochastic2019, nguyenFirstExitTime2019, zhangWhichAlgorithmicChoices2019,jinHowEscapeSaddle2017,simsekliTailIndexAnalysis2019} adds Gaussian noise to parameters. 
Randomness in training has been further exploited for uncertainty quantification. SWAG \cite{Maddox2019SimpleBaseline} and MultiSWAG \cite{Wilson2020ProbabilisticPerspective} use the SGD iterates to form a posterior approximation and MC Dropout \cite{galMCDropout2016} simply extends Dropout use to test time. 
Closest to our work is the recent study by \citet{Ghosh2025VariationalLearningFindsFlatter}, which investigates the edge of stability phenomenon in training procedures where a fixed amount of noise is injected at each optimization step. This setting can be interpreted as a proxy for variational learning in which only the mean parameters are optimized, together with an explicit entropy-based regularization term.
In contrast, our approach relies solely on the implicit regularization induced by the optimization dynamics: we do not include an explicit entropy or KL term, and we optimize the full variational distribution rather than only its mean. In this regime, because the noise itself is learned, it is \emph{a priori} unclear whether stochasticity persists throughout training or whether the dynamics collapse to a deterministic solution.

\section{Experiments}
\label{sec:experiments}

\paragraph{Experimental Setup.}
We train a set of ResNets on CIFAR10 and CIFAR100 via the cross-entropy loss using a standard and variational training setup \cite[\ie IBVI;][]{wengerVariationalDeepLearning2026}. 
For IBVI we use a variance-reduced version of the 1-sample approximation of the expected loss as proposed by \citet{Harrison2024VariationalBayesian}.
In all experiments we use a ResNet34 \cite{He2016DeepResidual} architecture initialized with pretrained weights except for the input and output layer.
The models are trained using SGD with momentum \(\momentum=0.9\) and a batch size of \(\batchsize=128\) for \(40\) epochs in single precision on an NVIDIA GH200 GPU.

We measure the accuracy, negative log-likelihood (NLL), and calibration error (ECE) on in- (ID) and out-of-distribution (OOD) test data. The OOD data are corrupted versions of the CIFAR10 and CIFAR100 test set, which are commonly used to assess robustness \cite{Hendrycks2019BenchmarkingNeural}. 
Throughout training we compute the sharpness of the cross-entropy loss at the current mean parameter \( \meanparam_t \) for the IBVI model and at the current weight \( \params_t \) for the standard training model. 
Results shown are averaged across five random seeds.
See \Cref{sec-supp:experiments} for more details on the experimental setup.

\paragraph{Sampling Reduces Sharpness of Global Minima.}

\begin{figure*}[t]
    \centering
    \begin{subfigure}[t]{0.33\textwidth}
        \centering
        \includegraphics[width=\textwidth]{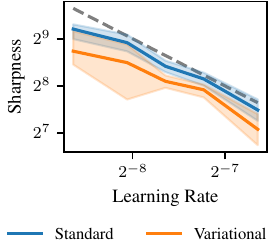}
        \caption{Sharpness across learning rates}
    \end{subfigure}
    \begin{subfigure}[t]{0.66\textwidth}
        \centering
        \includegraphics[width=\textwidth]{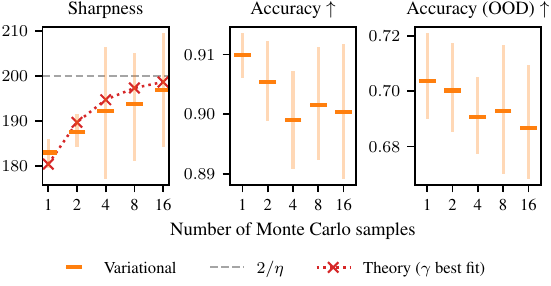}
        \caption{Results across number of samples \(M\)}
    \end{subfigure}
    \caption{\textit{Variational training reduces sharpness.} (a) Across learning rates on CIFAR-10, we observe that the noise introduced by sampling parameters during training on the expected loss (with \( M = 1 \) samples) consistently leads to less sharp minima, compared to standard training of neural networks. (b) Across the number of Monte Carlo samples $M$, we observe that smaller $M$ consistently leads to less sharp minima, with the sharpness tracking the theoretical prediction from \cref{eq:sharpness-threshold-general-loss}, $\frac{2}{\eta} \cdot \frac{1}{1 + 1/(M\gamma)}$, where $\gamma$ is fit by least squares jointly across all values of $M$. This translates into better in-distribution and out-of-distribution test accuracy, while larger $M$ consistently degrades performance on both.}
    \label{fig:lr-vs-sharpness}
\end{figure*}

To understand the effect of the additional regularization from a sampling approximation to the expected loss in IBVI, we measure the sharpness of the cross-entropy loss landscape at the current set of mean parameters throughout training. 
\Cref{fig:lr-vs-sharpness}(a) shows the average peak sharpness, corresponding to the point in training when the model reaches the edge of stability for standard training and unregularized variational training across a range of learning rates. 
As predicted by the edge of stability literature, the models trained in the standard way converge to a sharpness level that is approximately equal to \(2 / \lr\). 
Note that the small deviations from the theoretical limit could be explained by the fact that we train using SGD with momentum, rather than full batch gradient descent, which is typically assumed in the edge of stability literature.

As indicated by \Cref{thm:descent-general-loss}, variational training with \( M = 1 \) parameter sample consistently achieves smaller sharpness values.
Note that while theory suggests that the sharpness of the variational model should be approximately reduced by a factor of \( 2 \) (see \Cref{thm:descent-general-loss} with \( M = 1 \)), we observe a smaller gap in sharpness compared to standard training. 
A smaller sharpness gap suggests that, in our experiments, the signal-to-noise ratio (SNR) of the gradients with respect to sampled parameters is high towards the end of training, since our theoretical results show a positive dependence between the sharpness and SNR for a fixed learning rate.\footnote{A high SNR could be explained by the fact that we are only treating the input and output layer variationally, so the norm of the expected gradient should be significantly larger than the total variance in the gradients.}

To further understand the relationship between the number of Monte Carlo samples $M$ and the sharpness, we train IBVI models with different values of \( M \in \{1, 2, 4, 8, 16\} \). As shown in \Cref{fig:lr-vs-sharpness}(b, left), smaller values of $M$ consistently lead to less sharp minima, with the empirical sharpness tracking the theoretical prediction from \cref{eq:sharpness-threshold-general-loss} when the SNR $\gamma$ is fit by least squares across all values of $M$. Crucially, the sharpness saturates strictly below the standard edge-of-stability threshold $2/\eta$, and approaches it monotonically as $M$ grows---directly mirroring the limit $M \to \infty$ in which the gradient estimator becomes noiseless and the standard regime is recovered.

\paragraph{Lower Sharpness Improves Robustness.}

\begin{figure*}
    \centering
    \includegraphics[width=\textwidth]{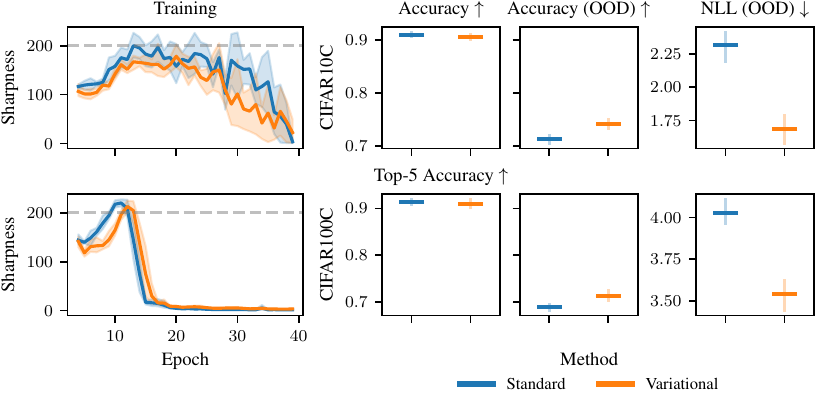}
    \caption{\textit{Sharpness, in-distribution, and out-of-distribution generalization.}
    We compare unregularized variational training to standard neural network training across multiple datasets and out-of-distribution tasks. As theory predicts, variational learning finds flatter minima, and the decay of sharpness toward the end of training is consistent with the implicit bias of gradient descent under cross-entropy loss (see \Cref{remark:celoss-sharpness-decay}). Despite optimizing the expected loss via parameter sampling, unregularized variational training matches standard training on in-distribution accuracy---indicating that the implicit bias does not degrade performance on the training distribution---while consistently outperforming it on out-of-distribution accuracy and negative log-likelihood.}
    \label{fig:experiments_main}
\end{figure*}

Minima with lower sharpness have been empirically associated with better generalization \cite{Keskar2017LargeBatchTraining,wuHowSGDSelects2018,Foret2021SharpnessAwareMinimization} and theoretically with higher regularity of the learned function \cite{Mulayoff2021ImplicitBias,Nacson2023ImplicitBias} and robustness to input perturbations \cite{Dherin2022WhyNeural}.
Based on our theoretical results in \Cref{sec:benefits-of-sampling} this suggests the additional noise induced by sampling parameters during variational training should generally result in more robust generalization. 
To assess this, we evaluate the trained ResNets on corrupted versions of CIFAR10 and CIFAR100 which are generated programmatically using a set of non-adversarial image corruptions such as increased brightness, blur, etc. from the test sets of the original dataset \cite{Hendrycks2019BenchmarkingNeural}.
As \Cref{fig:experiments_main} shows, standard training approximately reaches the edge of stability at a sharpness level of \( 2 / \lr \) (dashed line), whereas variational training with \( M = 1 \) samples consistently achieves smaller sharpness values.
This translates to significantly better OOD performance on the corrupted test sets both in terms of accuracy and in terms of uncertainty quantification (NLL, ECE). 

When training with $M \in \{1, 2, 4, 8, 16\}$ Monte Carlo samples on both clean and corrupted test data, as shown in \Cref{fig:lr-vs-sharpness}(b, middle, right), smaller $M$ yields higher in-distribution \emph{and} out-of-distribution test accuracy, with both metrics degrading monotonically as $M$ increases. This empirically supports the view that the gradient noise introduced by parameter sampling acts as an implicit regularizer biasing optimization toward flatter regions of the loss landscape, and suggests that small $M$---the cheapest regime---is not only sufficient but actively preferable for variational training.

\begin{remark}[Sharpness Decay for the Cross-Entropy Loss]
    \label{remark:celoss-sharpness-decay}
    For cross-entropy loss, the observed decay of sharpness toward zero after reaching the edge-of-stability at the end of training is a consequence of the implicit bias of gradient-based optimization. In particular, as training progresses, the norm of the parameters diverges \cite{soudryImplicitBiasGradient}. In this regime, the loss approaches zero, and the curvature of the loss landscape around the iterates vanishes \cite{cohenGradientDescentNeural2020,cohenCentralFlows2024}.
\end{remark}

\paragraph{Sampling Does Not Adversely Impact ID Performance.}

By sampling parameters from the variational distribution throughout training, variational training introduces additional stochasticity into the training objective in addition to the stochasticity arising from sampling minibatches.
As shown in \Cref{fig:experiments_main} and \Cref{fig:id-comparison} (in \Cref{sec-supp:additional-experimental-results}), this additional stochastic regularization does not materially impact in-distribution accuracy.
On the contrary, the variational method achieves better uncertainty quantification as measured in terms of NLL and calibration error (ECE), which is consistent with prior work on variational training without explicit regularization \cite{wengerVariationalDeepLearning2026}.

\section{Conclusion}
We studied the edge of stability phenomenon in variational deep learning without explicit prior regularization, in the regime where the expected loss is approximated by Monte Carlo sampling of the weights.
While sampling in variational learning is typically regarded as an unfortunate computational necessity, we showed that it is in fact a source of implicit regularization. 
Specifically, the maximal learning rates that guarantee descent, and consequently the sharpness at the edge of stability, are modified by the signal-to-noise ratio of the gradient estimates and the number of Monte Carlo samples, with fewer samples yielding flatter minima.
We rigorously characterized this phenomenon in the case of overparametrized linear regression, and extended the results to arbitrary smooth loss landscapes.
Finally, through several large-scale experiments on deep neural networks, we empirically verified the modified edge of stability and showed that the resulting bias toward flatter minima translates into improved out-of-distribution generalization. 
Notably, these gains come at no cost to in-distribution accuracy and at minimal computational overhead.

While we empirically demonstrated in large-scale experiments that sharpness is lower in the variational approach, the theoretical analysis is limited to smooth losses, does not account for stochastic gradient descent or momentum-based optimization methods, and is restricted to Gaussian variational families with a low-rank covariance factor. We expect similar behavior in more expressive families such as normalizing flows, but such analysis has not been conducted.
Our empirical findings provide a foundation for further evaluation across a broader range of architectures, datasets, and loss functions, and for exploring the influence of different variational estimators. Lastly, our results concern variational training rather than variational inference; the consequences of sampling-induced implicit regularization to the quality of the learned distribution are an interesting direction for the future work.

We believe there are multiple directions for extending the results presented in this manuscript. %
From a theoretical perspective, extending the results to non-smooth functions would be a substantial extension.
From a more practical perspective, we believe that the results of \Cref{thm:descent-mc-expected-loss} could be used to inform learning rate schedulers to potentially speed up variational training. Further directions also include comparing parameter-sampling noise with minibatch noise and dropout, and extending the analysis to variational inference and autoencoder-style objectives.

\subsubsection*{Acknowledgments}
JM, JW, BC and JPC are supported by the Gatsby Charitable Foundation (GAT3708), the Simons Foundation (542963), the NSF AI Institute for Artificial and Natural Intelligence (ARNI: NSF DBI 2229929) and the Kavli Foundation.
This work used the DeltaAI system at the National Center for Supercomputing Applications through allocations CIS250340 and CIS250292 from the Advanced Cyberinfrastructure Coordination Ecosystem: Services \& Support (ACCESS) program, which is supported by U.S. National Science Foundation grants \#2138259, \#2138286, \#2138307, \#2137603, and \#2138296.

\section*{Impact Statement}
This paper presents work whose goal is to advance the field of machine learning. 
There are many potential societal consequences of our work, none of which we feel must be specifically highlighted here.

{
    \setlength\bibitemsep{0.5\baselineskip}
	\printbibliography
}

\newpage

\newpage
\appendix
\onecolumn

\section{Proofs}
\myfavtheorem*

\begin{proof}
    Let \( \variationalparams = (\meanparam,\covfactorparam) \) and define the stochastic update
    \[
        \hat{d} = -\lr \nabla_{\variationalparams}\hat{\expectedloss}(\variationalparams),
        \qquad
        \lr=\mathrm{diag}(\lr_{\meanparam},\lr_{\covfactorparam}),
    \]
    where \( \nabla_{\variationalparams}\hat{\expectedloss}(\variationalparams) \) is obtained by averaging \(M\) Monte Carlo samples.
    Writing \(\hat{d}=(\hat{d}_{\meanparam},\hat{d}_{\covfactorparam})\), we have
    \begin{align*}
        \expval[z_1, \ldots, z_M]{\expectedloss (\variationalparams + \hat{d}) - \expectedloss (\variationalparams)}
        &=
        \expval[z_1, \ldots, z_M]{\nabla_\variationalparams \expectedloss (\variationalparams)^\top \hat{d} + \frac{1}{2} \hat{d}^\top \hessian[\variationalparams]{\expectedloss}{\variationalparams} \hat{d}} \\
        &= 
        \expval[z_1, \ldots, z_M]{\nabla_\meanparam \expectedloss (\meanparam)^\top \hat{d}_\meanparam + \frac{1}{2} \hat{d}_\meanparam^\top \hessian[\meanparam]{\expectedloss}{\meanparam} \hat{d}_\meanparam} \\
        &\quad \quad \quad +
        \expval[z_1, \ldots, z_M]{\nabla_\covfactorparam \expectedloss (\covfactorparam)^\top \hat{d}_\covfactorparam + \frac{1}{2} \hat{d}_\covfactorparam^\top \hessian[\covfactorparam]{\expectedloss}{\covfactorparam} \hat{d}_\covfactorparam},
    \end{align*}
    where the last equality follows from the block-diagonal structure of the Hessian for linear regression.

    We analyze the two contributions separately. Define the gradients of the expected loss with respect to the mean and covariance parameters as
    \begin{equation}\label{eq::linear_model_dmean}
        g := \nabla_{\meanparam}\expectedloss(\meanparam,\covfactorparam)
        = \mX^\top\big(\mX\meanparam-\vy\big),
    \end{equation}
    and
    \begin{equation}\label{eq::linear_model_dcov}
        G := \nabla_{\covfactorparam}\expectedloss(\meanparam,\covfactorparam)
        = \mX^\top \mX\,\covfactorparam.
    \end{equation}
    We begin with the mean update term. 
    \paragraph{Mean parameter update.}
    Note that \( \nabla_{\variationalparams}\hat{\expectedloss}(\variationalparams) \) is an unbiased estimator of
    \( \nabla_{\variationalparams}\expectedloss(\variationalparams) \).\footnote{For the mean parameter,
    \(
    \nabla_{\meanparam}\hat{\expectedloss}(\meanparam,\covfactorparam)
    =
    \mX^\top\big(\mX(\meanparam+\covfactorparam z)-\vy\big),
    \)
    and hence
    \(
    \expval{ \nabla_{\meanparam}\hat{\expectedloss}(\meanparam,\covfactorparam)}
    =
    \mX^\top(\mX\meanparam-\vy)
    =
    \nabla_{\meanparam}\expectedloss(\meanparam,\covfactorparam).
    \)
    An analogous calculation yields the claim for the covariance parameter.}
    Moreover, we have
    \[
        \hessian[\meanparam]{\expectedloss}{\meanparam} = \mX^\top \mX.
    \]
    Finally, writing the Monte Carlo gradient using \(M\) samples \(\{z_m\}_{m=1}^M\), we obtain
    \[
        \nabla_{\meanparam}\hat{\expectedloss}(\meanparam,\covfactorparam)
        =
        g + G\,\overline{z},
        \qquad
        \overline{z}:=\frac{1}{M}\sum_{m=1}^M z_m,
    \]
    and therefore the mean update direction can be written as
    \[
        \hat{d}_{\meanparam}
        =
        -\lr_{\meanparam}\big(g + G\,\overline{z}\big).
    \]
    
    We have
    \begin{align*}
        \Delta_\meanparam
        & :=
        \expval[z_1, \ldots, z_M]{\nabla_\meanparam \expectedloss (\meanparam)^\top \hat{d}_\meanparam + \frac{1}{2} \hat{d}_\meanparam^\top \hessian[\meanparam]{\expectedloss}{\meanparam} \hat{d}_\meanparam} \\ 
        &=
        - \lr_\meanparam \| g \|_2^2 + \frac{1}{2}  \expval[z_1, \ldots, z_M]{\hat{d}_\meanparam^\top \hessian[\meanparam]{\expectedloss}{\meanparam} \hat{d}_\meanparam} \\
        &=
        - \lr_\meanparam \| g \|_2^2 + \lr_\meanparam^2 \expval[z_1, \ldots, z_M]{(g + G \overline{z})^\top \mX^\top \mX (g + G \overline{z})} \\
        &=
        - \lr_\meanparam \| g \|_2^2 + \lr_\meanparam^2 \left ( g^\top \mX^\top \mX g + \frac{1}{M} \trace{G^\top \mX^\top \mX G} \right ) \\
        &=
        - \lr_\meanparam g^\top (\mId - \lr_\meanparam \mX^\top \mX) g + 4 \frac{\lr_\meanparam^2}{M} \trace{(\mX^\top \mX)^3 \covfactorparam^\top \covfactorparam} \\
        &= \lr_\meanparam^2 g^\top \mX^\top \mX g - \lr_\meanparam \| g \|_2^2 + 4 \frac{\lr_\meanparam^2}{M} \trace{(\mX^\top \mX)^3 \covfactorparam^\top \covfactorparam} \\
        &\leq
        \left [  \lr_\meanparam \left ( \lambda_{\max} (\mX^\top \mX) + \frac{4}{M} \frac{\trace{(\mX^\top \mX)^3 \covfactorparam^\top \covfactorparam}}{\| g \|_2^2}  \right ) - 1 \right ] \lr_\meanparam \| g \|_2^2.
    \end{align*}
    Next, by von Neumann's trace inequality we have that
    \[
        \trace{(\mX^\top \mX)^3 \covfactorparam^\top \covfactorparam} \leq \lambda_{\max} (\mX^\top \mX) \cdot \| \mX^\top \mX \covfactorparam \|_F^2. 
    \] 
    Thus, combining the above results, we have that
    \begin{equation}
        \Delta_\meanparam
        \leq
        \left [  \lr_\meanparam \lambda_{\max} (\mX^\top \mX) \left ( 1 + \frac{1}{M} \frac{\| \mX^\top \mX \covfactorparam \|_F^2}{\| \mX^\top (\mX \meanparam - \vy) \|_2^2}  \right ) - 1 \right ] \lr_\meanparam \| g \|_2^2.
    \end{equation}
    Hence, we have \( \Delta_\meanparam < 0 \) if
    \[
    \lr_\meanparam < \frac{1}{\lambda_\text{max} (\mX^\top \mX)} \left ( \frac{1}{1 + \frac{1}{M} \frac{ \| \mX^\top \mX \covfactorparam \|_F^2}{\| \mX^\top (\mX \meanparam - \vy) \|_2^2}} \right ).
    \]
    We continue by similiarly analyzing the second (covariance) term.
    \paragraph{Covariance parameter update.}
    Similar as before, let \( \overline{z} = \tfrac{1}{M} \sum_{m=1}^M z_m \). We have
    \begin{align*}
        \Delta_\covfactorparam
        & :=
        \expval[z_1, \ldots, z_M]{\nabla_\covfactorparam \expectedloss (\covfactorparam)^\top \hat{d}_\covfactorparam + \frac{1}{2} \hat{d}_\covfactorparam^\top \hessian[\covfactorparam]{\expectedloss}{\covfactorparam} \hat{d}_\covfactorparam} \\ 
        &=
        - \lr_\covfactorparam \| G \|_F^2 + \frac{1}{2} \lr_\covfactorparam^2 \expval[z_1, \ldots, z_M]{\hat{d}_\covfactorparam^\top \hessian[\covfactorparam]{\expectedloss}{\covfactorparam} \hat{d}_\covfactorparam} \\
        &=
        - \lr_\covfactorparam \| G \|_F^2 + \lr_\covfactorparam^2 \expval[z_1, \ldots, z_M]{\trace{(g \overline{z}^\top + \frac{G}{M} \sum_{m=1}^M z_m z_m^\top)^\top \mX^\top \mX (g \overline{z}^\top + \frac{G}{M} \sum_{m=1}^M z_m z_m^\top)}} \\
        &=
        - \lr_\covfactorparam \| G \|_F^2 + \lr_\covfactorparam^2 \Bigl [ \trace*{\expval[z_1, \ldots, z_M]{\overline{z}^\top \overline{z}} g^\top \mX^\top \mX g} + \frac{2}{M} \trace*{\sum_{m=1}^M \expval[z_1, \ldots, z_M]{z_m z_m^\top \overline{z}}g^\top \mX^\top \mX G} \\
        & \quad \hskip 4cm +
        \frac{1}{M^2} \trace*{ \sum_{m=1}^M \sum_{j=1}^M \expval[z_1, \ldots, z_M]{z_m z_m^\top z_j z_j^\top } G^\top \mX^\top \mX G} \Bigr ].
    \end{align*}
    Next, we compute the required expectations over standard normal random variables. We have
    \begin{itemize}
        \item \(  \expval[z_1, \ldots, z_M]{\overline{z}^\top \overline{z}}  = \trace*{ \frac{1}{M} \mId_{R \times R}} = \frac{R}{M}, \)
        \item \( \expval[z_1, \ldots, z_M]{z_m z_m^\top \overline{z}} = \frac{1}{M} \expval[z_m]{z_m z_m^\top z_m} = 0, \)
        \item \( \expval[z_1, \ldots, z_M]{z_m z_m^\top z_j z_j^\top} = \begin{cases}
            (R + 2) \mId_{R \times R}, & m = j, \\
            \mId_{R \times R}, & m \neq j.
        \end{cases} \)
    \end{itemize}
    Plugging these results back into the expression for \( \Delta_\covfactorparam \) we have
    \begin{align*}
        \Delta_\covfactorparam
        &=
        - \lr_\covfactorparam \| G \|_F^2 + \lr_\covfactorparam^2 \left ( \frac{R}{M} g^\top \mX^\top \mX g + \frac{R + M + 1}{M} \trace{G^\top \mX^\top \mX G} \right  ) \\
        &=
        \lr_\covfactorparam \left ( \lr_\covfactorparam \left ( \frac{R}{M} g^\top \mX^\top \mX g + \frac{R + M + 1}{M} \trace{G^\top \mX^\top \mX G} \right ) -  \| G \|_F^2  \right ) \\
        &\leq
        \left [  \lr_\covfactorparam \lambda_{\max} (\mX^\top \mX) \left ( \frac{R}{M} \| g \|_2^2 + \frac{R + M + 1}{M} \| G \|_F^2 \right ) -  \| G \|_F^2 \right ] \lr_\covfactorparam \\
        &=
        \left [  \lr_\covfactorparam  \lambda_{\max} (\mX^\top \mX) \left ( \frac{R}{M} \frac{\| g \|_2^2}{\| G \|_F^2}  + \frac{R + M + 1}{M} \right ) -  1 \right ] \lr_\covfactorparam \| G \|_F^2 \\
        &=
        \left [  \lr_\covfactorparam  \lambda_{\max} (\mX^\top \mX) \left ( \frac{R}{M} \frac{\| \mX^\top (\mX \meanparam - \vy) \|_2^2}{\| \mX^\top \mX \covfactorparam \|_F^2}  + \frac{R + M + 1}{M} \right ) -  1 \right ] \lr_\covfactorparam \| G \|_F^2.
    \end{align*}
    Hence, we have \( \Delta_\covfactorparam < 0 \) if
    \[
        \lr_\covfactorparam
        <
        \frac{1}{\lambda_\text{max} (\mX^\top \mX)} \left ( \frac{1}{ \frac{R + M + 1}{M} + \frac{R}{M} \frac{\| \mX^\top (\mX \meanparam - \vy) \|_2^2}{\| \mX^\top \mX \covfactorparam \|_F^2} } \right ).
    \]
    This finishes the proof.
\end{proof}

\lemmaJointDescent*

\begin{proof}
    Recall the definition of \( \Delta_\meanparam \) and \( \Delta_\covfactorparam \) from the proof of \Cref{thm:descent-mc-expected-loss}. We are interested in characterizing the joint conditions on \( \lr_\meanparam \) and \( \lr_\covfactorparam \) such that \( \Delta_\meanparam + \Delta_\covfactorparam < 0. \) By plugging in the upper bounds for \( \Delta_\meanparam \) and \( \Delta_\covfactorparam \) from the proof of \Cref{thm:descent-mc-expected-loss} we have
    \begin{align*}
        \Delta_\meanparam + \Delta_\covfactorparam
        &\leq
        \left [  \lr_\meanparam \lambda_{\max} (\mX^\top \mX) \left ( 1 + \frac{1}{M} \frac{\| \mX^\top \mX \covfactorparam \|_F^2}{\| \mX^\top (\mX \meanparam - \vy) \|_2^2}  \right ) - 1 \right ] \lr_\meanparam \| g \|_2^2 \\
        &\quad +
        \left [  \lr_\covfactorparam  \lambda_{\max} (\mX^\top \mX) \left ( \frac{R}{M} \frac{\| \mX^\top (\mX \meanparam - \vy) \|_2^2}{\| \mX^\top \mX \covfactorparam \|_F^2}  + \frac{R + M + 1}{M} \right ) -  1 \right ] \lr_\covfactorparam \| G \|_F^2 \\
        &= 
        \| g \|^2_2 \left ( a_\mu \lr_\meanparam^2 - \lr_\meanparam \right ) + \| G \|_F^2 \left ( a_S \lr_\covfactorparam^2 - \lr_\covfactorparam \right ),
    \end{align*}
    where we defined
    \[
        a_\mu := \lambda_{\max} (\mX^\top \mX) \left ( 1 + \frac{1}{M} \frac{\| \mX^\top \mX \covfactorparam \|_F^2}{\| \mX^\top (\mX \meanparam - \vy) \|_2^2}  \right ),
    \]
    \[
        a_S := \lambda_{\max} (\mX^\top \mX) \left ( \frac{R}{M} \frac{\| \mX^\top (\mX \meanparam - \vy) \|_2^2}{\| \mX^\top \mX \covfactorparam \|_F^2}  + \frac{R + M + 1}{M} \right ).
    \]
    Hence we have \( \Delta_\meanparam + \Delta_\covfactorparam < 0 \) if
    \[
        \| g \|^2_2 \left ( a_\mu \lr_\meanparam^2 - \lr_\meanparam \right ) + \| G \|_F^2 \left ( a_S \lr_\covfactorparam^2 - \lr_\covfactorparam \right ) < 0,
    \]
    which corresponds to elliptical conditions on the learning rates \( \lr_\meanparam \) and \( \lr_\covfactorparam. \)
\end{proof}

\theoremGeneralLoss*

\begin{proof}
    We begin the proof by showing that the mapping \( \meanparam \mapsto \expectedloss(\meanparam, \covfactorparam) \) is again \( \beta \)-smooth for any \( \covfactorparam \).
    To that end, using differentiation under the integral sign (justified e.g.\ by dominated convergence since \(\loss\) is \(C^2\) with uniformly bounded Hessian), we have
    \begin{equation}
    \begin{split}
        \nabla^2_{\meanparam}\expectedloss(\meanparam,\covfactorparam)
        &= \nabla^2_{\meanparam} \expval[z]{\loss(\meanparam+\covfactorparam z)} \\
        &= \expval[z]{\nabla^2_{\omega}\loss(\omega)\big|_{\omega=\meanparam+\covfactorparam z}}.
    \end{split}
    \end{equation}
    Therefore, by Jensen's inequality for the operator norm,
    \begin{equation}
    \begin{split}
        \big\|\nabla^2_{\meanparam}\expectedloss(\meanparam,\covfactorparam)\big\|_2
        &= \Big\| \expval[z] {\nabla^2_{\omega}\loss(\meanparam+\covfactorparam z)}\Big\|_2 \\
        &\leq \expval[z]{\big\|\nabla^2_{\omega}\loss(\meanparam+\covfactorparam z)\big\|_2} \\
        &\leq \expval[z]{\beta} = \beta.
    \end{split}
    \end{equation}
    Hence \( \meanparam \mapsto \expectedloss(\meanparam,\covfactorparam)\) is \( \beta \)-smooth for any fixed \(\covfactorparam\).

    Let \( \hat{d}_{\meanparam} = - \lr_{\meanparam} \nabla_{\meanparam}\hat{\expectedloss}(\meanparam,\covfactorparam) \), where \( \nabla_{\meanparam}\hat{\expectedloss} \) is the Monte Carlo estimator obtained by averaging \(M\) i.i.d.\ samples.
    Since \( \meanparam \mapsto \expectedloss(\meanparam,\covfactorparam) \) is \( \beta \)-smooth for any fixed \(\covfactorparam\), the descent inequality yields
    \begin{align*}
    \expectedloss(\meanparam+\hat{d}_{\meanparam},\covfactorparam)
    &\leq
    \expectedloss(\meanparam,\covfactorparam)
    +
    \nabla_{\meanparam}\expectedloss(\meanparam,\covfactorparam)^\top \hat{d}_{\meanparam}
    +
    \frac{\beta}{2} \|\hat{d}_{\meanparam}\|_2^2 .
    \end{align*}
    Taking expectation over the Monte Carlo randomness \((z_1,\ldots,z_M)\) gives
    \begin{align*}
    \Delta_\meanparam :=
    \expval[z_1,\ldots,z_M]{
    \expectedloss(\meanparam+\hat{d}_{\meanparam},\covfactorparam)
    -
    \expectedloss(\meanparam,\covfactorparam)}
    &\leq
    \expval[z_1,\ldots,z_M]{\nabla_{\meanparam}\expectedloss(\meanparam,\covfactorparam)^\top \hat{d}_{\meanparam}}
    +
    \frac{\beta}{2}
    \expval[z_1,\ldots,z_M]{\|\hat{d}_{\meanparam}\|_2^2}.
    \end{align*}
    
    Denote \( g := \nabla_\meanparam \expectedloss (\meanparam) \). We obtain
    \begin{align*}
        \Delta_{\meanparam}
        &=
        -\lr_{\meanparam} \|g\|_2^2
        +\frac{\beta}{2} \lr_{\meanparam}^2
        \expval[z_1,\ldots,z_M]{\big\|\nabla_{\meanparam}\hat{\expectedloss}(\meanparam,\covfactorparam)\big\|_2^2} \\
        &=
        -\lr_{\meanparam} \|g\|_2^2
        +\frac{\beta}{2} \lr_{\meanparam}^2
        \Big(
        \|g\|_2^2
        +
        \trace*{
        \cov[z_1,\ldots,z_M]{\nabla_{\meanparam}\hat{\expectedloss}(\meanparam,\covfactorparam)}
        }
        \Big) \\
        &=
        \lr_{\meanparam}\|g\|_2^2 \left(
        \frac{\beta \lr_{\meanparam}}{2} \left(
        1+
        \frac{
        \trace*{ \cov[z_1,\ldots,z_M]{\nabla_{\meanparam}\hat{\expectedloss}(\meanparam,\covfactorparam)}}
        }{
        \|g\|_2^2
        }
        \right)
        -1
        \right).
    \end{align*}
    If \(\nabla_{\meanparam}\hat{\expectedloss}\) is formed by averaging \(M\) i.i.d.\ samples
    \(\nabla_{\meanparam}\loss(\params)\) with \(\params \sim \mathcal N(\meanparam,\covfactorparam\covfactorparam^\top)\),
    then
    \[
        \cov[z_1,\ldots,z_M]{\nabla_{\meanparam}\hat{\expectedloss}}
        =
        \frac{1}{M}
        \cov[\params][\mathcal N(\meanparam,\covfactorparam\covfactorparam^\top)]{\nabla \loss(\params)},
    \]
    and therefore
    \begin{align*}
        \Delta_{\meanparam}
        &=
        \lr_{\meanparam} \|g\|_2^2
        \left(
        \frac{\beta \lr_{\meanparam}}{2}
        \left(
        1+
        \frac{1}{M}
        \frac{
        \trace*{ \cov[\params][\mathcal N(\meanparam,\covfactorparam\covfactorparam^\top)]{\nabla \loss(\params)} }
        }{
        \|g\|_2^2
        }
        \right)
        -1
        \right) \\
        &=
        \lr_{\meanparam} \|g\|_2^2
        \left(
        \frac{\beta \lr_{\meanparam}}{2}
        \left(
        1+
        \frac{1}{M}
        \frac{
        \trace*{ \cov[\params][\mathcal N(\meanparam,\covfactorparam\covfactorparam^\top)]{\nabla \loss(\params)} }
        }{
        \big\|\expval[\params][\mathcal N(\meanparam,\covfactorparam\covfactorparam^\top)]{\nabla \loss(\params)}\big\|_2^2
        }
        \right)
        -1
        \right).
    \end{align*}

    Hence, we have \( \Delta_\meanparam < 0 \) if
    \[
        \lr_\meanparam < \frac{2}{\beta} 
        \left ( \frac{1}{ 1 + \frac{1}{M}  \frac{\trace*{ \cov[\params][\variationalpdf_\variationalparams]{\nabla \loss (\params) }}}{\| \expval[\params][\variationalpdf_\variationalparams]{\nabla \loss (\params)} \|_2 ^2}} \right ).
    \]
\end{proof}

\section{Additional Insights}

\subsection{A view from the function space}
\begin{figure}[h!]
    \centering
    \includegraphics[width=0.99\textwidth]{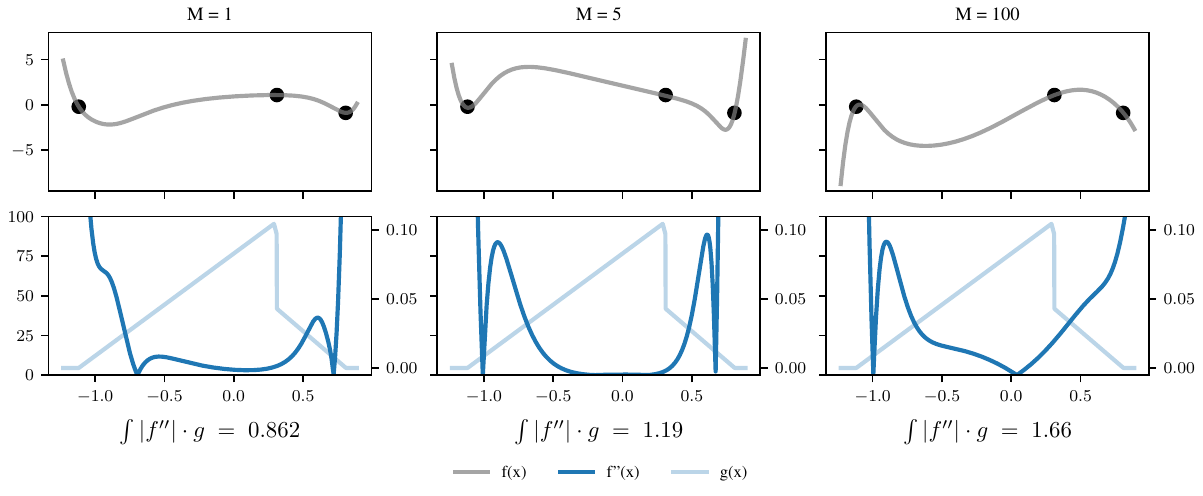}
    \caption{\textit{A more detailed view of sampling effects on function smoothness.} This figure provides additional detail corresponding to \Cref*{fig:figure1}. We use a weighting function inspired by the approach of \citet{Mulayoff2021ImplicitBias}, which studies the implicit bias of stable minima from a function-space perspective. The first row reproduces the results shown in \Cref*{fig:figure1}. The second row displays the absolute value of second derivative \( |f''| \) of the learned function together with the data-weighting function \( g \), highlighting how different sampling strategies affect local curvature. The associated smoothness measure is computed via the seminorm \( \int |f''(x)|  g(x) \, \mathrm{d}x \), which emphasizes regions of the input space weighted by the sampling distribution. As the number of Monte Carlo samples \( M \) increases, the learned functions become less regular. In these experiments, we use \( N = 3 \) training data points and a one-layer ReLU IBVI model with hidden size \( 100 \), trained using full gradient descent for \( 20{,}000 \) epochs with learning rate \( \lr = 0.05 \).
    }

    \label{fig:smooth-vs-M}
\end{figure}

\subsection{Non-Asymptotic Error Analysis}

\begin{restatable}[Non-Asymptotic Error of SGD]{theorem}{nonasymptoticErrorSGD}
    \label{thm:nonasymptotic-error-sgd}
    Let \(f_\params(\vx) = \vx^\top \params\) be a linear model. Define a prior \(p(\params) = \gaussianpdf{\params}{\meanparam_0}{\covfactorparam_0\covfactorparam_0^\tr}\) and assume noise-free observations \(y(\cdot) = f_\params(\cdot)\) for \(\params \sim p(\params)\). Further, define a variational distribution \smash{\(\variationalpdf_{\variationalparams}(\params) = \gaussianpdf{\params}{\meanparam}{\covfactorparam\covfactorparam^\tr}\)} with \(\variationalparams = (\meanparam,\covfactorparam)\) such that \(\meanparam \in \R^{\paramspacedim}\) and \smash{\(\covfactorparam \in \R^{\paramspacedim \times \covrank}\)} where \(\covrank \leq \paramspacedim\). Assume the expected loss is given by \(\expectedloss(\variationalparams) = \expval[\variationalpdf_\variationalparams(\params)]{\frac{1}{2} \norm*{\traintargets - \traininputs \params}_2^2}\) and let \(\variationalparams(\idxoptimstep) = (\meanparam(\idxoptimstep), \covfactorparam(\idxoptimstep))\) be the variational parameters at step \(\idxoptimstep\) of (stochastic) gradient descent with learning rate sequence \((\lr_\idxoptimstep)_\idxoptimstep\), initialized at \(\variationalparams(0) = (\meanparam_0, \covfactorparam_0)\). Then the expected squared error of the mean prediction
    \begin{equation}
        \label{eqn:non-asymptotic-error-sgd}
        \expval[\big(\substack{\traintargets \\ y_\symboltestdata}\big)]{\big(y_\symboltestdata - f_{\meanparam(\textcolor{MPLred}{\idxoptimstep})}(\testpoint)\big)^2} = \var[\params \sim \variationalpdf_{\variationalparams(\textcolor{MPLred}{\idxoptimstep})}]{f_\params(\testpoint)}
    \end{equation}
    at any test point \(\testpoint \in \R^\paramspacedim\).
    In other words, assuming the training and test data are drawn from the prior predictive, the predictive error of \(f_{\meanparam(\idxoptimstep)}(\cdot)\) at any optimization step \(\idxoptimstep\) is \emph{exactly} quantified by the predictive uncertainty of the variational distribution.
    
    Further, if the learning rate \(\lr_\idxoptimstep \leq \frac{1}{\lambda_{\max}(\traininputs_\idxoptimstep^\tr \traininputs_\idxoptimstep)}\) for all steps \(\idxoptimstep\), then
    \begin{equation}
        \label{eqn:uncertainty-decreases-monotonically}
        \trace{\cov[\params \sim \variationalpdf_{\variationalparams(\idxoptimstep + 1)}]{\params}} \leq \trace{\cov[\params \sim \variationalpdf_{\variationalparams(\idxoptimstep)}]{\params}},
    \end{equation}
    \ie uncertainty about the parameters decreases monotonically during optimization.
\end{restatable}

\begin{proof}
    See Theorem~S3 in \citet{wengerVariationalDeepLearning2026}.
\end{proof}

\subsection{Evolution of the Variational Covariance During Training}
\label{sec-supp:covariance-evolution}

\begin{figure}[h]
    \centering
    \includegraphics[width=\textwidth]{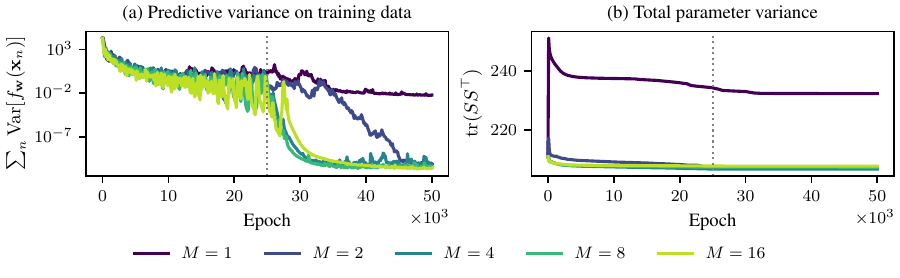}
    \caption{\textit{The variational distribution collapses only in the directions spanned by the training data.}
    Evolution of the variational covariance on the synthetic regression problem of \Cref{fig:figure1} for different numbers of parameter samples \( M \).
    (a) The total predictive variance on the training inputs decays by several orders of magnitude, and the collapse happens later for smaller \( M \).
    (b) In contrast, the total variance \( \trace{\covfactorparam \covfactorparam^\tr} \) of the variational parameters stays close to its value at initialization, so the variance in the directions orthogonal to the training data is retained.
    The dotted line marks the end of the learning rate warm-up.}
    \label{fig:covariance-evolution}
\end{figure}

Since the variational distribution is trained without any explicit regularization towards a prior, a natural question is whether, and when, it collapses during training. To investigate this, we track the covariance \( \covfactorparam \covfactorparam^\tr \) of the variational layer over the course of training (see \Cref{fig:covariance-evolution}) on the synthetic regression problem of \Cref{fig:figure1} with \( N = 3 \) training data points. 

We use a one-hidden-layer SiLU network with hidden size \( 100 \), in which only the output layer is variational with a full-rank covariance factor, and train it for \( 50{,}000 \) epochs of full-batch gradient descent with learning rate \( \lr = 0.02 \), linearly warmed up over the first half of training, using \( M \in \{1, 2, 4, 8, 16\} \) parameter samples. Since we use full-batch gradients, parameter sampling is the only source of gradient noise. Every \( 100 \) epochs we record the total variance \( \trace{\covfactorparam \covfactorparam^\tr} \) of the variational parameters, as well as the total predictive variance on the training inputs \( \sum_{n=1}^{N} \var[\params \sim \variationalpdf_{\variationalparams}]{f_\params(\vx_n)} \), which measures the variance in the directions spanned by the learned representation of the training data. 

We observe the same qualitative behavior for all values of \( M \). The predictive variance on the training inputs decays by several orders of magnitude over the course of training, \ie the variational distribution collapses in the directions spanned by the learned features of the training data, consistent with the analysis of \citet{wengerVariationalDeepLearning2026} for overparametrized linear models. 
This collapse occurs later for smaller \( M \): for \( M \geq 4 \) it happens shortly after the learning rate reaches its final value, for \( M = 2 \) it is noticeably delayed, and for \( M = 1 \) the predictive variance on the training data is still decaying at the end of training. We attribute this to the higher variance of the gradient estimate for small \( M \).

\clearpage

\subsection{Additional Experimental Results} \label{sec-supp:additional-experimental-results}
\begin{figure}[h]
    \includegraphics[width=\columnwidth]{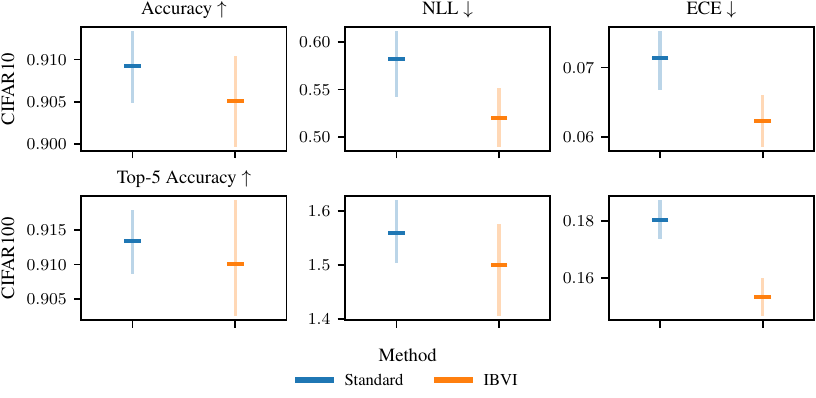}
    \caption{\textit{Implicit bias does not hurt in-distribution performance.} We observe that variational training achieves performance comparable to standard training on in-distribution metrics across multiple datasets. Despite optimizing the expected loss via parameter sampling, the model maintains in-distribution accuracy, indicating that implicit bias does not degrade performance on the training distribution. In terms of uncertainty quantification, IBVI improves upon the standard approach to training.}
    \label{fig:id-comparison}
\end{figure}

\section{Experimental Details}
\label{sec-supp:experiments}

Our experiments build on \texttt{inferno}, the open-source library implementing unregularized variational training \cite{wengerVariationalDeepLearning2026}, available at \url{https://github.com/inferno-ml/inferno}, which we extend with sharpness estimation.

In all of our experiments we used the following datasets and metrics.

\subsection{Datasets}
\label{sec-supp:datasets}

\begin{table}[H]
    \caption{\emph{Benchmark datasets used in our experiments.} All corrupted datasets are only intended for evaluation and thus only have
    test sets consisting of 15 different corruptions of the original test set.}
    \centering
    \begin{tabular}{lrrrrc}
    \toprule
    Dataset & \(\numtraindata\) & \(\numtestdata\) & \(\inputdim\) & \(\numclasses\) & Train / Validation Split\\
    \midrule
    CIFAR-10 \cite{Krizhevsky2009LearningMultiple} & 50\,000 & 10\,000 & 3 \(\times\) 32 \(\times\) 32 & 10 & (0.9, 0.1)\\
    CIFAR-100 \cite{Krizhevsky2009LearningMultiple} & 50\,000 & 10\,000 & 3 \(\times\) 32 \(\times\) 32 & 100 & (0.9, 0.1)\\
    \midrule
    CIFAR-10-C \cite{Hendrycks2019BenchmarkingNeural} & - & 150\,000 & 3 \(\times\) 32 \(\times\) 32 & 10 & -\\
    CIFAR-100-C \cite{Hendrycks2019BenchmarkingNeural} & - & 150\,000 & 3 \(\times\) 32 \(\times\) 32 & 100 & -\\
    \bottomrule
\end{tabular}

\end{table}

\subsection{Metrics}
\label{sec-supp:metrics}

\paragraph{Accuracy}
The (top-k) accuracy is defined as
\begin{equation}
    \operatorname{Accuracy}_k = \frac{1}{\numtraindata} \sum_{\idxdata=1}^{\numtraindata} 1_{(y_\idxdata \in \argmax_{I \subset [C]: \abs{I} = k} f_w(\vx_\idxdata))}.
\end{equation}

\paragraph{Negative Log-Likelihood (NLL)}
The (normalized) negative log likelihood for classification is given by
\begin{equation}
    \operatorname{NLL} = - \frac{1}{\numtraindata}\sum_{\idxdata=1}^{\numtraindata} \log \operatorname{softmax}(f_w(\vx_\idxdata))_{\hat{y}_\idxdata},
\end{equation}
where \(\hat{y}_\idxdata = \argmax f_w(\vx_\idxdata)\) is the model's predicted class on a given datapoint \(\vx_\idxdata\).

\paragraph{Expected Calibration Error (ECE)}
The expected calibration error measures how well a model is calibrated, \ie how closely the predicted class probability matches the accuracy of the model.
Assume the predicted probabilities of the model on the test set are binned into a given binning of the unit interval. 
Compute the accuracy \(a_j\) and average predicted probability \(\hat{p}_j\) of each bin, then the expected calibration error is given by 
\begin{equation}
    \operatorname{ECE} = \sum_{j=1}^{J} b_j \abs{a_j - \hat{p}_j},
\end{equation}
where \(b_j\) is the fraction of datapoints in bin \(j \in \{1, \dots, J\}\).

\paragraph{Sharpness}
The sharpness is defined as the maximum eigenvalue of the Hessian of the loss 
\begin{equation}
    \operatorname{sharpness}(f_{\params}) = \lambda_\text{max} (\hessian[\params]{\loss}{\params})
\end{equation}
with respect to the model's parameters \(\params \in \R^\paramspacedim\). In practice, we estimate the sharpness by averaging over the sharpness per batch of training data. The largest eigenvalue of the Hessian on a batch is computed via a warm-started Lanczos process, which avoids the memory overhead of instantiating the Hessian, since it only relies on Hessian-vector products.

\end{document}